\documentclass{article}
\usepackage{ijcai17}

\usepackage{times}

\usepackage{times}
\usepackage{helvet}
\usepackage{courier}

\usepackage[utf8]{inputenc} 
\usepackage[T1]{fontenc}    
\usepackage{url}            
\usepackage{booktabs}       
\usepackage{amsfonts}       
\usepackage{microtype}      
\usepackage{latexsym}
\usepackage{graphicx}
\usepackage{algorithm}
\usepackage{algorithmic}
\usepackage{amsmath}
\usepackage[normalem]{ulem}
\usepackage{color}
\usepackage{caption,booktabs,array}\usepackage{multirow}
\usepackage{subfig}
\usepackage[font=small]{caption}

\title{Exemplar-centered Supervised Shallow Parametric Data Embedding}
\author{
Martin Renqiang Min\\
NEC Labs America \\
Princeton, NJ 08540\\
renqiang@nec-labs.com
\And
Hongyu Guo\\
National Research Council Canada\\ 
Ottawa, ON K1A 0R6\\
hongyu.guo@nrc-cnrc.gc.ca
\And
Dongjin Song\\
NEC Labs America \\
Princeton, NJ 08540\\
dosong@nec-labs.com
}

\begin{document}

\maketitle

\begin{abstract}
Metric learning methods for dimensionality reduction in combination  with k-Nearest Neighbors (kNN) have been extensively deployed in many classification, data embedding, and information retrieval applications. However, most of these approaches involve pairwise training data comparisons, and thus have quadratic computational complexity with respect to the size of training set, preventing them from scaling to fairly big datasets. Moreover, during testing, comparing test data against all the training data points is also expensive in terms of both computational cost and resources required. Furthermore, previous metrics are either too constrained or too expressive to be well learned. To effectively solve these issues, we present an exemplar-centered supervised shallow parametric data embedding model, using a Maximally Collapsing Metric Learning (MCML) objective. Our strategy learns a shallow high-order parametric embedding function and compares training/test data only with learned or precomputed exemplars, resulting in a cost function with linear computational complexity for both training and testing. We also empirically demonstrate, using several benchmark datasets, that for classification in two-dimensional embedding space, our approach not only gains speedup of kNN by hundreds of times, but also outperforms state-of-the-art supervised embedding approaches.
\end{abstract}
\section{Introduction}
Given the class information of training data, metric learning methods for dimensionality reduction and data visualization essentially learn a linear or nonlinear transformation from a high-dimensional input feature space to a low-dimensional embedding space, aiming at  increasing the similarity between pairwise data points from the same class while decreasing the similarity between pairwise data points from different classes in the embedding space. These methods in combination with kNN have been widely used in many applications including computer vision, information retrieval, and bioinformatics. Recent surveys on metric learning can be found in \cite{Survey1,Bellet}. However, most of these approaches, including the popular Maximally Collapsing Metric Learning (MCML)~\cite{MCML2006}, Neighborhood Component Analysis (NCA)~\cite{NCA2005}, and Large-Margin Nearest Neighbor (LMNN)~\cite{Weinberger:2009}, need to model neighborhood structures by comparing pairwise training data points either for learning parameters or for constructing target neighborhoods in the input feature space, which results in quadratic computational complexity requiring careful tuning and heuristics to get approximate solutions in practice and thus limits the methods' scalability. Moreover, during testing, kNN is often employed to compare each test data point against all training data points in the input feature or embedding space, which is also expensive in terms of both computational cost and resources required. In addition, a lot of previous methods, e.g., MCML, on one extreme, focus on learning a Mahalanobis metric that is equivalent to learning a linear feature transformation matrix and thus incapable of achieving the goal of collapsing classes. On the other extreme, nonlinear metric learning methods based on deep neural networks such as dt-MCML and dt-NCA~\cite{MinMYBZ10} are powerful but very hard to learn and require complicated procedures such as tuning network architectures and tuning many hyperparameters. For data embedding and visualization purposes, most users are reluctant to go through these complicated procedures, which explains why dt-MCML and dt-NCA were not widely used although they are much more powerful than simpler MCML, NCA, and LMNN.

To address the aforementioned issues of previous metric learning methods for dimensionality reduction and data visualization, in this paper, we present an exemplar-centered supervised shallow parametric data embedding model based on a Maximally Collapsing Metric Learning objective and Student $t$-distributions. Our model learns a shallow high-order parametric embedding function that is as powerful as a deep neural network but much easier to learn. Moreover, during training, our model avoids pairwise training data comparisons and compares training data only with some jointly learned exemplars or precomputed exemplars from supervised k-means centers, resulting in an objective function with linear computational complexity with respect to the size of training set. In addition, during testing, our model only compares each test data point against a very small number of exemplars. As a result, our model in combination with kNN accelerates kNN using high-dimensional input features by hundreds of times owing to the benefits of both dimensionality reduction and sample size reduction, and achieves much better performance. Even surprisingly, in terms of both accuracy and testing speed, our shallow model based on pre-computed exemplars significantly outperforms state-of-the-art deep embedding method dt-MCML. We also empirically observe that, using a very small number of randomly sampled exemplars from training data, our model can also achieve competitive classification performance. We call our proposed model exemplar-centered High Order Parametric Embedding (en-HOPE).
Our contributions in this paper are summarized as follows: (1) We propose a salable metric learning strategy for data embedding with an objective function of linear computational complexity, avoiding pairwise training data comparisons; (2) Our method compares test data only with a small number of exemplars and gains speedup of kNN by hundreds of times; (3) Our approach learns a simple shallow high-order parametric embedding function, beating state-of-the-art embedding models on several benchmark datasets in term of both speed and accuracy.

\section{Related Work}\label{sec:related}
Metric learning methods and their applications have been comprehensively surveyed in~\cite{Survey1,Bellet}. Among them, our proposed method en-HOPE is closely related to the ones that can be used for dimensionality reduction and data visualization, including MCML~\cite{MCML2006}, NCA~\cite{NCA2005}, LMNN~\cite{Weinberger:2009}, nonlinear LMNN~\cite{NLMNN}, and their deep learning extensions such as dt-MCML~\cite{MinMYBZ10}, dt-NCA~\cite{MinMYBZ10}, and DNet-kNN~\cite{MinICDM}. en-HOPE is also related to neighborhood-modeling dimensionality reduction methods such as LPP~\cite{He2003}, t-SNE~\cite{van2008visualizing}, its parametric implementation SNE-encoder~\cite{MinThesis} and deep parametric implementation pt-SNE~\cite{Maaten09}. The objective functions of all these related methods have at least quadratic computational complexity with respect to the size of training set due to pairwise training data comparisons required for either loss evaluations or target neighborhood constructions. Our work is also closely related to the RVML method~\cite{DBLP:conf/nips/PerrotH15}, which suffers scalability issues as MCML does.

en-HOPE is closely related to a recent sample compression method called Stochastic Neighbor Compression (SNC)~\cite{SNC2014} for accelerating kNN classification in a high-dimensional input feature space. SNC learns a set of high-dimensional exemplars by optimizing a modified objective function of NCA. en-HOPE differs from SNC in several aspects: First, their objective functions are different; Second, en-HOPE learns a nonlinear metric based on a shallow model for dimensionality reduction and data visualization, but SNC does not have such capabilities; Third, en-HOPE does not necessarily learn exemplars, instead, which can be precomputed. We will compare en-HOPE to SNC in the experiments to evaluate the compression ability of en-HOPE, however, the focus of en-HOPE is for data embedding and visualization but not for sample compression in a high-dimensional space. 

en-HOPE learns a shallow parametric embedding function by considering high-order feature interactions. High-order feature interactions have been studied for learning Boltzmann Machines, autoencoders, structured outputs, feature selections, and biological sequence classification~\cite{DBLP:conf/iccv/Memisevic11,DBLP:conf/aistats/MinNCG14,MinPSB14,DBLP:conf/cvpr/RanzatoH10,DBLP:journals/jmlr/RanzatoKH10,DBLP:journals/corr/GuoZM15,Min2014kdd,MinBio2015}. To the best of our knowledge, our work here is the first successful one to model input high-order feature interactions for  supervised data embedding and exemplar learning.

\section{Method}\label{sec:method}
In this section, we introduce MCML and dt-MCML at first. Then we describe our shallow parametric embedding function based on high-order feature interactions. Finally, we present our scalable model en-HOPE.
\subsection{A Shallow Parametric Embedding Model for Maximally Collapsing Metric Learning}\label{sec:sup}
Given a set of data points $\mathcal{D} = \{{\mathbf x}^{(i)}, L^{(i)}: i = 1,\ldots, n\}$, where ${\mathbf x}^{(i)} \in {\mathbb R}^H$ is the input feature vector, $L^{(i)} \in \{1, \ldots, c\}$ is the class label of a labeled data point, and $c$ is the total number of classes. MCML learns a Mahalanobis distance metric to collapse all data points in the same class to a single point and push data points from different classes infinitely farther apart. Learning a Mahalanobis distance metric can be thought of as learning a linear feature transformation ${\mathbf y} = f({\mathbf x}) = {\mathbf A}{\mathbf x}$  from the high-dimensional input feature space to a low-dimensional latent embedding space, where ${\mathbf A} \in {\mathbb R}^{h \times H}$, and $h < H$.  For data visualization, we often set $h = 2$.

MCML assumes, $q_{j|i}$, the probability of each data point $i$ chooses every other data point $j$ as its nearest neighbor in the latent embedding space follows a Gaussian distribution,
\begin{equation}\label{asymmqg}
q_{j|i} = \frac{\exp(-d_{ij})}{\sum_{k: k \neq i} {\exp(-d_{ik})}}, \quad q_{i|i} = 0.
\end{equation} 
and 
\begin{equation}\label{dij}
d_{ij} = ||f({\mathbf x}^{(i)}) -  f({\mathbf x}^{(j)})||^2. 
\end{equation}

To maximally collapse classes, MCML minimizes the sum of the Kullback-Leibler divergence between the  conditional  probabilities  $q_{j|i}$ computed in the embedding space and  the ``ground-truth'' probabilities $p_{j|i}$ calculated based  on  the  class labels of training data. Specifically, $p_{j|i} \propto 1$ iff $L^{(i)} = L^{(j)}$ and $p_{j|i} = 0$ iff $L^{(i)} \neq L^{(j)}$.
Formally, the objective function of the MCML is as follows:
\begin{equation}\label{obj}
\small
\ell = \sum_{ij: i \neq j} p_{j|i}\log \frac{p_{j|i}}{q_{j|i}} \propto -\sum_{ij: i \neq j} [L^{(i)} = L^{(j)}]\log q_{j|i} + const,
\end{equation}
where $[\cdot]$ is an indicator function. 

However, learning a Mahalanobis metric requires solving a positive semidefinite programming problem, which is computationally prohibitive and prevents MCML from scaling to a fairly big dataset. Moreover, a linear feature transformation is very constrained and makes it impossible for MCML to achieve its goal of collapsing classes. dt-MCML extends MCML in two aspects: (1) it learns a powerful deep neural network to parameterize the feature transformation function ${\mathbf y} = f({\mathbf x})$; (2) it uses a symmetric heavy-tailed $t$-distribution to compute $q_{j|i}$ for supervised embedding due to its capabilities of reducing overfitting, creating tight clusters, increasing class separation, and easing gradient optimization. Formally, this stochastic neighborhood metric first centers a $t$-distribution over ${\mathbf y}^{(i)}$, and then computes the density of ${\mathbf y}^{(j)}$ under the distribution as follows.
\begin{eqnarray}
q_{j|i} & = & \frac{(1 + d_{ij})^{-1}} {\sum_{kl:k \neq l}  (1 + d_{kl})^{-1}}, \quad q_{ii} = 0, \label{eqn:symmq}
\end{eqnarray}

Although dt-MCML based on a deep neural network has a powerful nonlinear feature transformation, parameter learning is hard and requires complicated procedures such as tuning network architectures and tuning many hyperparameters. Most users who are only interested in data embedding and visualization  are reluctant to go through these complicated procedures. Here we propose to use high-order feature interactions, which often capture structural knowledge of input data, to learn a shallow parametric embedding model instead of a deep model. The shallow model is much easier to train and does not have many hyperparameters. In the following, the shallow high-order parametric embedding function will be presented. We expand each input feature vector ${\mathbf x}$ to have an additional component of $1$ for absorbing bias terms, that is, ${\mathbf x}^{\prime} = [{\mathbf x};   1]$, where ${\mathbf x}^{\prime} \in {\mathbb R}^{H+1}$. The $O$-order feature interaction is the product of all possible $O$ features $\{x_{i_1}\times \ldots \times  x_{i_t} \times \ldots \times x_{i_O}\}$ where,  $t \in \{1, \ldots, O\}$, and $\{i_1, \ldots, i_t, \ldots, i_O\} \in \{1, \ldots, H\}$. Ideally, we want to use each $O$-order feature interaction as a coordinate and then learn a linear transformation to map all these high-order feature interactions to a low-dimensional embedding space. However, it's very expensive to enumerate all possible $O$-order feature interactions. For example, if $H = 1000, O = 3$, we must deal with a $10^9$-dimensional vector of high-order features. We approximate a Sigmoid-transformed high-order feature mapping ${\mathbf y} = f({\mathbf x})$ by constrained tensor factorization as follows (derivations omitted due to space constraint), 
\begin{equation}
\label{shopemap}
y_s = \sum_{k=1}^m V_{sk} \sigma(\sum_{f=1}^F W_{fk}({\mathbf C_f}^T {\mathbf x}^{\prime})^O + b_k),
\end{equation}
where $b_k$ is a bias term, ${\mathbf C}  \in {\mathbb R}^{(H+1) \times F}$ is a factorization matrix, ${\mathbf C}_f$ is the $f$-th column of ${\mathbf C}$, ${\mathbf W}\in {\mathbb R}^{F\times m}$ and ${\mathbf V}\in {\mathbb R}^{h\times m}$ are projection matrices, $y_s$ is the $s$-th component of ${\mathbf y}$, $F$ is the number of factors, $m$ is the number of high-order hidden units, and $\sigma (x) = \frac{1}{1 + e^{-x}}$.  Because the last component of $\mathbf{x}^\prime$ is 1 for absorbing bias terms, the full polynomial expansion of $({\mathbf C_f}^T {\mathbf x}^\prime)^O$ essentially captures all orders of input feature interactions up to order $O$. Empirically, we find that $O=2$ works best for all datasets we have and set $O=2$ for all our experiments. The hyperparameters $F$ and $m$ are set by users.

Combining Equation~\ref{obj}, Equation~\ref{eqn:symmq} and the feature transformation function in Equation~\ref{shopemap} leads to a method called High Order Parametric Embedding (HOPE). As MCML and dt-MCML, the objective function of HOPE involves comparing pairwise training data and thus has quadratic computational complexity with respect to the sample size. The parameters of HOPE are learned by Conjugate Gradient Descent. 

\subsection{en-HOPE for Data Embedding and Fast kNN Classification}
Building upon HOPE for data embedding and visualization described earlier,  we present two related approaches to implement en-HOPE, resulting in an objective function with linear computational complexity with respect to the size of training set.  The underlying intuition is that, instead of comparing pairwise training data points, we compare training data only with a small number of exemplars in the training set to achieve the goal of collapsing classes, collapsing all training data to the points defined by exemplars. In the first approach, we simply precompute the exemplars by supervised k-means and only update the parameters of the embedding function during training. In the second approach, we simultaneously learn exemplars and embedding parameters during training.  During testing, fast kNN classification can be efficiently performed in the embedding space against a small number of exemplars especially when the dataset is huge. 

Given the same dataset $\mathcal{D}$ with formal descriptions as introduced in Section~\ref{sec:sup}, 
we aim to obtain $z$ exemplars from the whole dataset with their designated class labels uniformly sampled from the training set to account for data label distributions, where $z$ is a user-specified free parameter and $z << n$. We denote these exemplars by $\{{\mathbf e}^{(j)}:  j = 1, \ldots, z\}$. In the first approach, we perform k-means on the training data to identify the same number of exemplars as in the sampling step for each class (please note that k-means often converges within a dozen iterations and shows linear computational cost in practice).  Then we minimize the following objective function to learn high-order embedding parameters ${\mathbf \Theta}$ while keeping the exemplars $\{{\mathbf e}^{(j)}\}$ fixed,
\begin{eqnarray}\label{exobj}
&\min_{}^{} \ell({\mathbf \Theta}, \{{\mathbf e}^{(j)}\})  =  \sum_{i=1}^{n}\sum_{j=1}^{z} p_{j|i}\log \frac{p_{j|i}}{q_{j|i}} \nonumber \\ 
& \propto  -\sum_{i=1}^{n} \sum_{j=1}^{z} [L^{(i)} = L^{(j)}]\log q_{j|i} + const
\end{eqnarray}
where $i$ indexes training data points, $j$ indexes exemplars, ${\mathbf \Theta}$ denotes the high-order embedding parameters $\{\{b_k\}_{k=1}^m, \mathbf{C }, \mathbf{W}, \mathbf{V}\}$ in Equation~\ref{shopemap}, $p_{j|i}$ is calculated in the same way as in the previous description, 
but $q_{j|i}$ is calculated with respect to exemplars,
\begin{eqnarray}
q_{j|i} & = & \frac{(1 + d_{ij})^{-1}} {\sum_{i=1}^n\sum_{k=1}^z (1 + d_{ik})^{-1}},\\
d_{ij} & = & ||f({\mathbf x}^{(i)}) - f({\mathbf e}^{(j)}) ||^2,
\end{eqnarray}
where $f(\cdot)$ denotes the high-order embedding function as described in Equation~\ref{shopemap}. Note that unlike the probability distribution in Equation~\ref{eqn:symmq},  $q_{j|i}$ here is computed only using the pairwise distances between training data points and exemplars. This small modification has significant benefits. Because $z << n$, compared to the quadratic computational complexity with respect to $n$ of Equation~\ref{obj}, the objective function in Equation~\ref{exobj} has a linear computational complexity with respect to $n$. In the second approach, we jointly learn the high-order embedding parameters ${\mathbf \Theta}$ and the exemplars $\{{\mathbf e}^{(j)}\}$ simultaneously by optimizing the objective function in Equation~\ref{exobj}. The derivative of the above objective function with respect to exemplar ${\mathbf e}^{(j)}$ is as follows,
\begin{eqnarray}
\frac{\partial \ell({\mathbf \Theta}, \{{\mathbf e}^{(j)}\})} {\partial {\mathbf e}^{(j)}} & = & 
\sum_{i=1}^n 2 (1 + d_{ij})^{-1}(p_{j|i} - q_{j|i} )  \nonumber \\
&&(f({\mathbf e}^{(j)}) 
 - f({\mathbf x}^{(i)}))
\frac{\partial    f({\mathbf e}^{(j)})}{\partial {\mathbf e}^{(j)}}
\end{eqnarray}
In both approaches to implementing en-HOPE, all the model parameters are learned using Conjugate Gradient Descent. We call the first approach en-HOPE (k-means exemplars) and the second approach en-HOPE (learned exemplars).
\begin{table*}[ht]
  \centering
  \caption{Error rates (\%) obtained by 5NN on the 2-dimensional representations produced by different dimensionality reduction methods on the MNIST, USPS, and 20Newsgroups datasets. Due to the nonscalability issue of the original MCML, it fails to run on the MNIST dataset. The results demonstrate the effectiveness of the shallow high-order parametric embedding.}
	\scalebox{0.95}{    
 \begin{tabular}{|lc|lc|lc|lc|lc|lc|}\hline
\multicolumn{4}{|c|}{MINIST}&\multicolumn{4}{c}{USPS}&\multicolumn{4}{|c|}{20 Newsgroups}\\ \hline
\multicolumn{2}{|c|}{Linear Methods}&\multicolumn{2}{c|}{Non-Linear Methods} &\multicolumn{2}{c|}{Linear Methods}&\multicolumn{2}{c|}{Non-Linear Methods} &\multicolumn{2}{c|}{Linear Methods}&\multicolumn{2}{c|}{Non-Linear Methods} \\ \hline
LPP&47.20&pt-SNE & 9.90 & LPP &34.77 &pt-SNE & 17.90 & LPP &24.64 &pt-SNE & 28.90\\
NCA&45.91&dt-NCA & 3.48 & NCA&37.17&dt-NCA & 5.11 & NCA& 30.84 &dt-NCA & 25.85\\
MCML$^+$&35.67&dt-MCML& 3.35 & MCML&44.60&dt-MCML & 4.07 & MCML &26.65 & dt-MCML & 21.10\\
LMNN&56.28& & & LMNN& 48.40&& & LMNN& 29.15 & &  \\
\hline
 &  &HOPE & \textbf{3.20} &   &  &HOPE& \textbf{3.03} &  & &HOPE & \textbf{20.05}\\

\hline
\end{tabular}}

  \label{tab:accuracy:mnist}
\end{table*}

\begin{table}[t]
  \centering
   \caption{Error rates (\%) by kNN on the 2-dimensional representations produced by HOPE and en-HOPE and other methods on top of VGG features of MNIST data. The kNN error rate in the original 512-dimensional space generated by VGG is 0.62, which is comparable to the kNN performance on the 2-dimensional representations produced by HOPE and en-HOPE.
  }\label{tab:vggknn}
 \begin{tabular}{|c|c|}\hline
 Methods & Error Rates \\
 \hline 
VGG + LMNN & 1.75 \\
\hline
VGG+ NCA & 1.83\\
\hline 
VGG + MCML$^+$  & 0.80\\
\hline
VGG + HOPE & \textbf{0.65} \\
\hline
\hline
VGG + LMNN (s-kmeans) & 2.22\\
\hline
VGG + NCA (s-kmeans) & 2.18 \\
\hline
VGG + en-HOPE (10 k-means exemplars) & 0.67\\
\hline
VGG + en-HOPE (10 learned exemplars) & 0.66\\
\hline
VGG + en-HOPE (20 k-means exemplars) & \textbf{0.64}\\
\hline
VGG + en-HOPE (20 learned exemplars) & 0.68\\
\hline
VGG + en-HOPE (10 random exemplars) & 0.68\\  
\hline
\end{tabular}
\end{table}

\section{Experiments}\label{sec:experiment}
In this section, we evaluate the effectiveness of HOPE and en-HOPE by comparing them against several baseline methods based upon three datasets, \textit{i.e.}, MNIST, USPS, and 20 Newsgroups. The MNIST dataset contains 60,000 training and 10,000 test gray-level 784-dimensional images. The USPS data set contains 11,000 256-pixel gray-level images, with 8,000 for training and 3,000 for test. The 20 Newsgroups dataset is a collection of 16,242 newsgroup documents among which we use 15,000 for training and the rest for test as in~\cite{Maaten09}.

To evaluate whether our proposed shallow high-order parametric embedding function is powerful enough, we first compare HOPE with four linear metric learning methods, including LPP, LMNN, NCA, and MCML, and three deep learning methods without convolutions, including a deep unsupervised model pt-SNE, as well as two deep supervised models, \textit{i.e.}, dt-NCA and dt-MCML. To make computational procedures and tuning procedures for data visualization simpler, none of these models was pre-trained using any unsupervised learning strategy, although HOPE, en-HOPE, dt-NCA, and dt-MCML could all be pre-trained by autoencoders or variants of Restricted Boltzmann Machines~\cite{MinMYBZ10,MinBio2015}. 

We set the number of exemplars used to 10 and 20 in all our experiments. When 10 exemplars are used, $k = 1$ for kNN, otherwise, $k = 5$. We used $10\%$ of training data as validation set to tune the number of factors ($F$), the number of high-order units ($m$), and batch size. For HOPE and en-HOPE, we set $F=800$ and $m=400$ for all the datasets used. In practice, we find that the feature interaction order $O=2$ often works best for all applications. The parameters for all baseline methods were carefully tuned to achieve the best results.

\subsection{Classification Performance of High-order Parametric Embedding}
Table~\ref{tab:accuracy:mnist} presents the test error rates of 5-nearest neighbor classifier on 2-dimensional embedding generated by HOPE and some baseline methods. The error rate is calculated by the number of misclassified test data points divided by the total number of test data points. 
We chose 2D as in pt-SNE because we can effectively visualize and intuitively understand the quality of the constructed embeddings as will be presented and discussed later in this section. 
The results in Table~\ref{tab:accuracy:mnist} indicate that HOPE 
significantly outperforms its linear and nonlinear competitors on three datasets. Due to the nonscalability issue of the original MCML, it fails to run on the MNIST dataset. We implemented an improved version of MCML called MCML$^+$ by directly learning a linear feature transformation matrix based on conjugate gradient descent.

Promisingly, 
results in Table~\ref{tab:accuracy:mnist}  suggest that 
our shallow model HOPE even outperforms deep embedding models based on deep neural networks, in terms of accuracy obtained on the 2-dimensional embedding for visualization. For example, on MNIST, the error rate (3.20\%) of HOPE  is lower than the ones of the pt-SNE, dt-NCA, and dt-MCML methods. These results clearly demonstrate the representational efficiency and power of supervised shallow models with high-order feature interactions. 

To further confirm the representation power of HOPE, we extracted the 512-dimensional features of MNIST digits below the softmax layer learned by a well-known deep convolutional architecture VGG~\cite{Simonyan2015}, which currently holds the-state-of-the-art classification performance through a softmax layer on MNIST.  Next, we ran HOPE based on these features to generate 2D embedding. As is shown in the top part of Table~\ref{tab:vggknn}, VGG+HOPE can achieve an error of 0.65\%. In contrast, NCA and LMNN on top of VGG, respectively, produces test error rate of 1.83\% and 1.75\%. This error rate of HOPE  represents the historically low test error rate in two-dimensional space on MNIST, which implies that even on top of a powerful deep convolutional network, modeling explicit high-order feature interactions can further improve accuracy and outperform all other models without feature interactions.

\subsection{Experimental Results for Different Methods with Exemplar Learning}

\renewcommand{\arraystretch}{0.8}
\begin{table}[t]
  \centering
   \caption{Error rates (\%) obtained by kNN on the two-dimensional representations created by different testing methods with different exemplar learning methods on, respectively, MNIST (top), USPS (middle), and 20 Newsgroups(bottom).\vspace{2mm}
  }\label{tab:accuracy:exemplar}
  	\scalebox{0.8593}{
 \begin{tabular}{>{\quad}lc|lc}\hline
\multicolumn{2}{c|}{s-kmeans+methods}&\multicolumn{2}{c}{en-HOPE} \\ \hline
 LPP&45.13&    en-HOPE (10 k-means exemplars) &  2.86\\
 NCA&50.67& en-HOPE (10 learned exemplars) & 2.80\\
 LMNN&59.67& en-HOPE (20 k-means exemplars) & 2.72\\
 pt-SNE& 18.86& en-HOPE (20 learned exemplars) & \textbf{2.66}\\
 dt-MCML& 3.17 & en-HOPE (10 random exemplars) & 3.19\\
\hline\hline
 LPP&33.23& en-HOPE (10 k-means exemplars) &  2.96\\
 NCA&35.13& en-HOPE (10 learned exemplars) & \textbf{2.67}\\
 LMNN&59.67& en-HOPE (20 k-means exemplars) & 2.83\\
 pt-SNE& 29.47& en-HOPE (20 learned exemplars) & 3.03\\
 dt-MCML& 4.27 & en-HOPE (10 random exemplars) & 3.10\\ 
\hline\hline
 LPP&33.09& en-HOPE (10 k-means exemplars) &  \textbf{18.27}\\
 NCA&36.71& en-HOPE (10 learned exemplars) & 18.84\\
 LMNN&38.24& en-HOPE (20 k-means exemplars) & 19.64 \\
 pt-SNE&33.17&   en-HOPE (20 learned exemplars) & 18.44\\
 dt-MCML& 21.90 & en-HOPE (10 random exemplars) & 18.84\\ 
\hline
\end{tabular}}
\end{table}\vspace{2mm}

In this section, we evaluate the performance of en-HOPE for data embedding, data visualization, and fast kNN classification. Table~\ref{tab:accuracy:exemplar} presents the classification error rates of kNN on 2-dimensional embeddings generated by en-HOPE with the two proposed exemplar learning.  Exemplar-based en-HOPE consistently achieves better performance than the ones of HOPE in Table~\ref{tab:accuracy:mnist}. To construct stronger baselines, we run supervised k-means to get exemplars and train each baseline method independently. During testing, we only use these k-means centers for comparisons with test data. We call these experiments "s-kmeans+methods".  Please note that "s-kmeans+methods" heuristics have objective functions with quadratic computational complexity as the original baseline methods and thus are not scalable to big datasets. To test whether en-HOPE can indeed effectively collapse classes, we also randomly select data points from each class as fixed exemplars and then learn the high-order embedding function of en-HOPE. The results in Table~\ref{tab:accuracy:exemplar}  suggest the following: when coupled with exemplars, en-HOPE significantly outperforms other baseline methods including the deep embedding models; even with randomly sampled exemplars, for example, one exemplar per class on MNIST and USPS, en-HOPE with an objective function of linear computational complexity can still achieve very competitive performance compared to baseline methods, demonstrating the effectiveness of our proposed shallow high-order model coupled with exemplars for collapsing classes. The bottom part of Table~\ref{tab:vggknn}
again verifies the additional gain of our shallow high-order model en-HOPE on top of an established deep convolutional neural network.
\subsubsection{Two-dimensional Data Embedding Visualization}
\begin{figure*}[t]
      \subfloat[MCML$^+$\label{MNIST_MCML_embed}]{%
      \includegraphics[width=0.3738\textwidth]{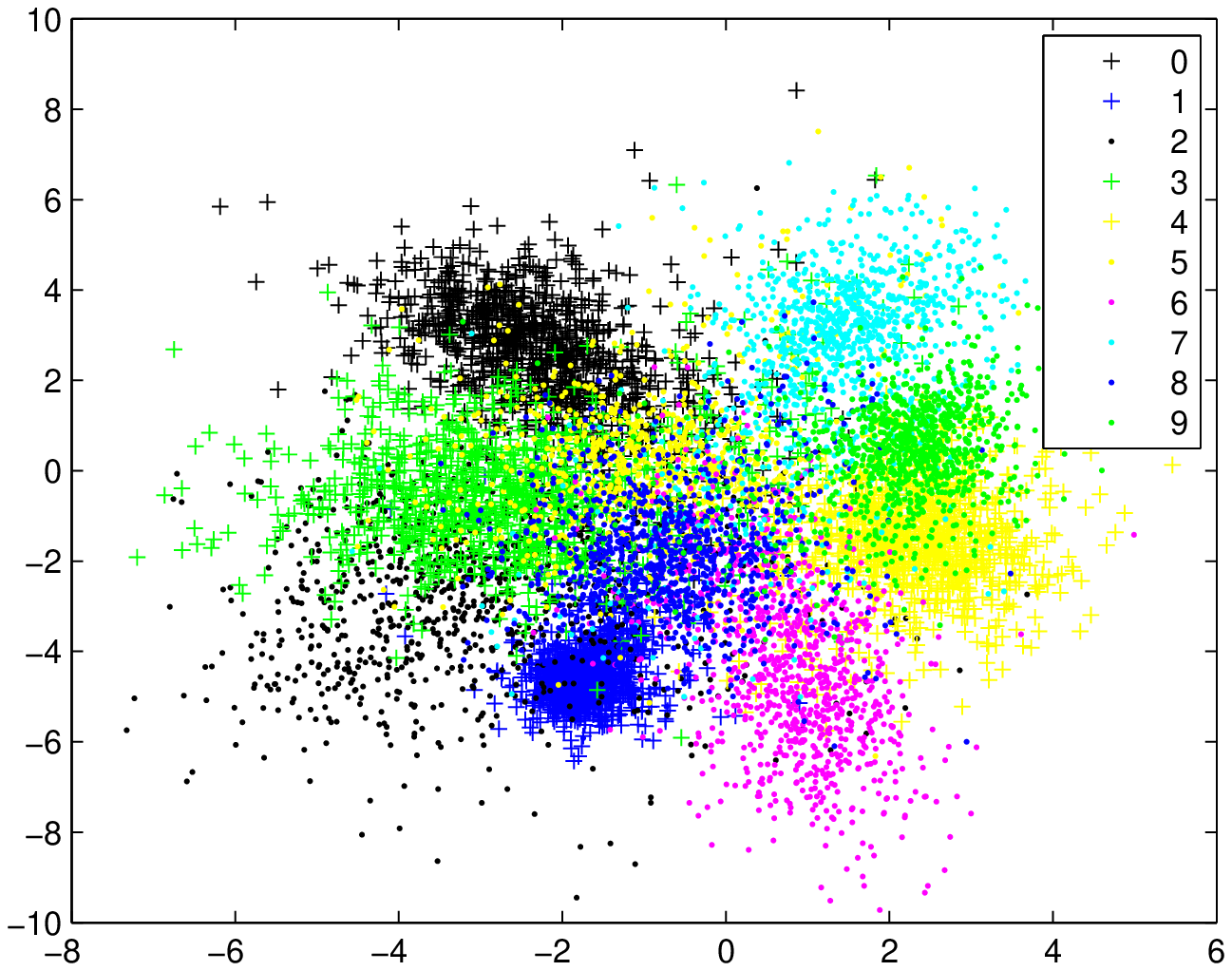}
      }
        \subfloat[dt-MCML\label{MNIST_dtMCML_embed}]{%
      \includegraphics[width=0.3738\textwidth]{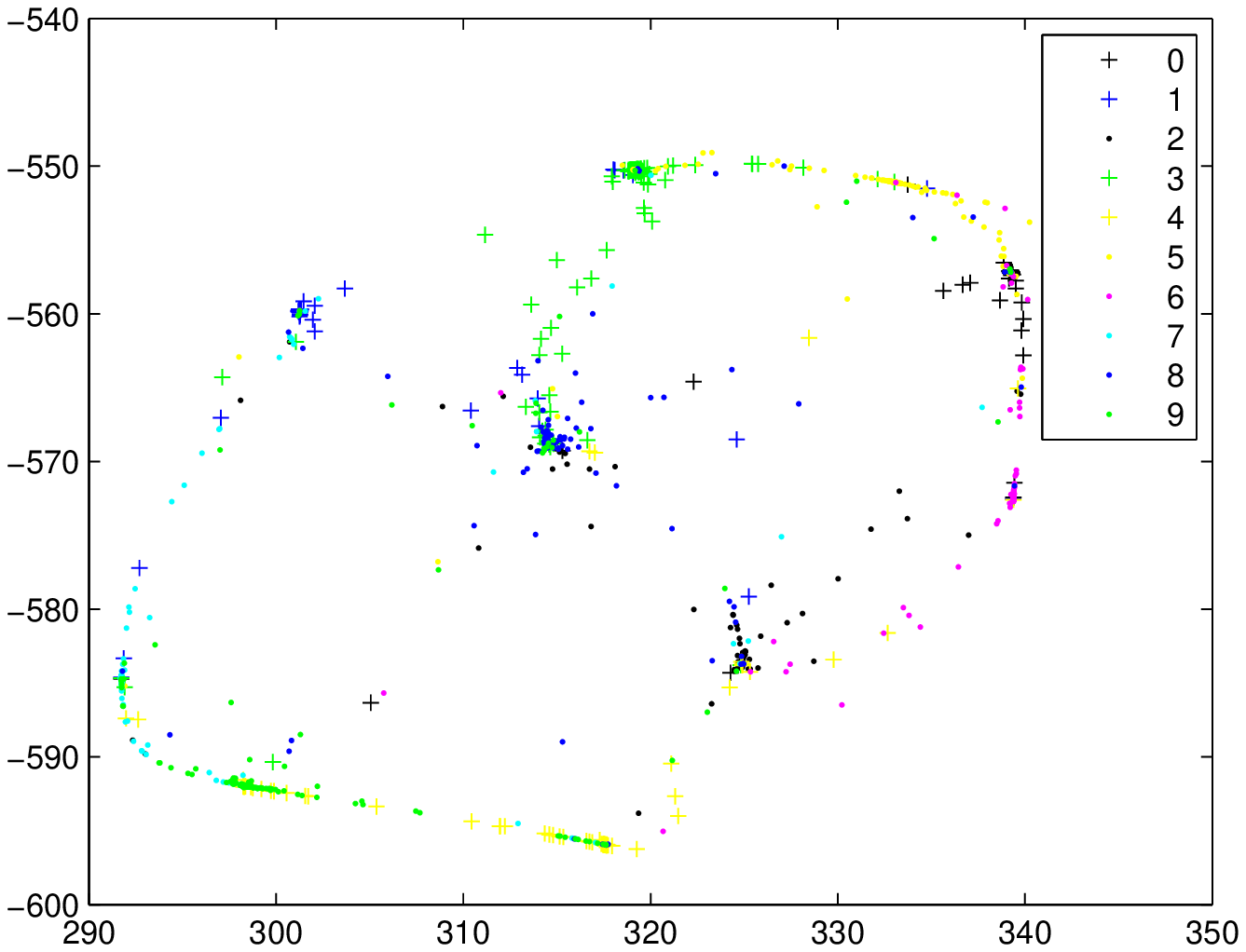}
    }
     \hfill
       \subfloat[en-HOPE with 20 learned exemplars\label{MNISTEx20learned}]{%
      \includegraphics[width=0.3738\textwidth]{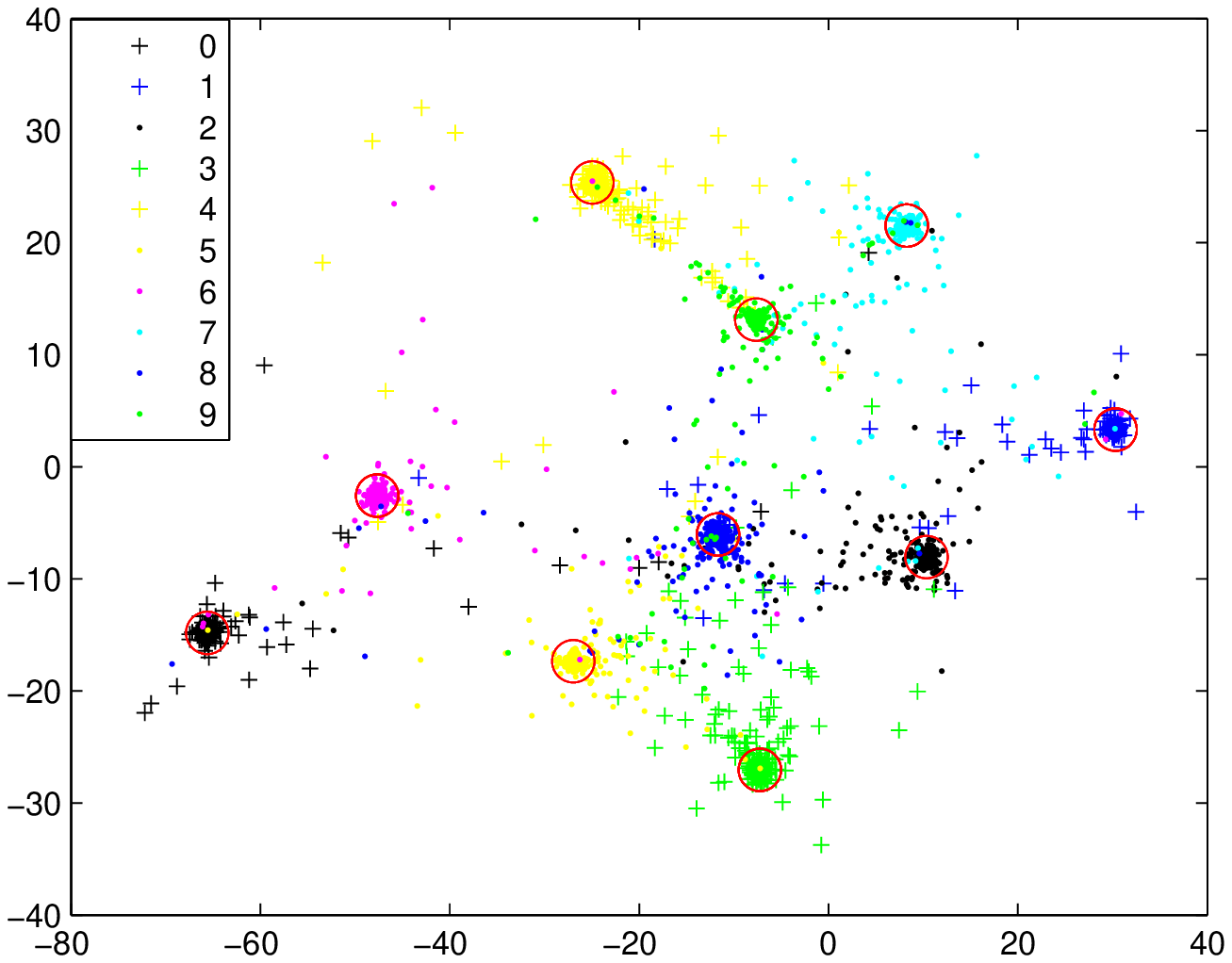}
    }
     \subfloat[en-HOPE with 10 learned exemplars \label{MNISTEx10learned}]{%
      \includegraphics[width=0.3738\textwidth]{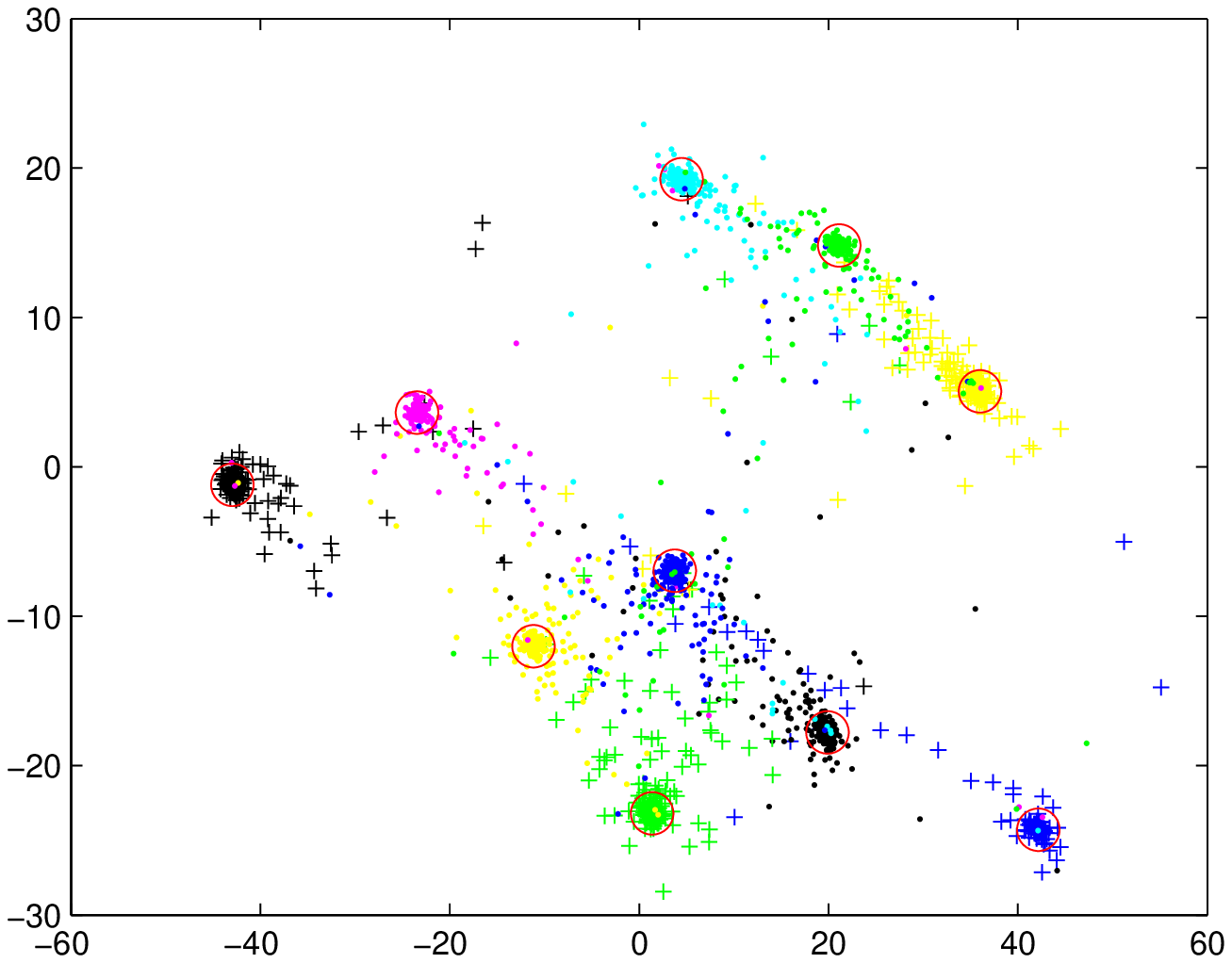}
    }
  \centering\vspace{1mm}
  \caption{2-dimensional embeddings of 10000 MNIST test data points constructed by
  MCML$^+$, dt-MCML,  en-HOPE (20 learned exemplars), and en-HOPE (10 learned exemplars); the red empty circles are the learned exemplars.}
  \label{fig:visualMNIST}
  \end{figure*}

Figure~\ref{fig:visualMNIST} shows the test data embeddings of MINST by different methods. These embeddings were constructed by, respectively, MCML$^+$, dt-MCML, en-HOPE with 20 learned exemplars, and en-HOPE with 10 learned exemplars. The 20 learned exemplars overlap in the two-dimensional space. en-HOPE produced  the  best visualization,  collapsed  all  the  data  points  in  the  same  class  close  to  each  other,  and generated large separations between class clusters. Furthermore, the embeddings of the learned exemplars created during training (depicted as red empty circles in subfigure (c) and (d)) are  located almost at the centers of all the clusters.

 \vspace{2mm}
  
\subsubsection{Computational Efficiency of en-HOPE for Sample Compression}
\begin{table}[t]
  \centering
   \caption{Observed computational speedup of en-HOPE with 20 learned exemplars over standard kNN on different datasets.
  }\label{tab:speedup}
 \begin{tabular}{|c|c|c|c|}\hline
 Datasets & MNIST & USPS & 20 Newsgroups \\ 
 \hline
 Speedup (times)& 463 $\times$ & 28 $\times$ & 101$\times$\\
\hline
Error rates of en- & & & \\
HOPE in 2D space& 2.66 & 3.03 & 18.44\\
\hline 
Error rates of kNN & & & \\
in high-D space& 3.05 & 4.77 & 25.12\\
\hline
\end{tabular}
\end{table}


en-HOPE speeds up computational efficiency of fast information retrieval such as kNN classification used in the above experiments by 
hundreds of times. Table~\ref{tab:speedup} shows the experimentally observed computational speedup of en-HOPE over standard kNN on our desktop with Intel Xeon 2.60GHz CPU and 48GB memory on different datasets. The test error rates by kNN in high-dimensional feature space are much worse than the ones produced by en-HOPE even in a much lower feature dimension, i.e., the two-dimensional latent space.  In detail, on our desktop, for classifying 10000 MNIST test data, standard kNN takes 124.97 seconds, 
but our method en-HOPE with 20 learned exemplars only takes 0.24 seconds including the time for computing the two-dimensional embedding of test data. In other words, our method en-HOPE has 
463 times speedup over standard kNN along with much better classification performance. This computational speedup will be more pronounced on massive datasets.

\subsubsection{Comparisons of en-HOPE with SNC and dt-MCML}
Stochastic neighbor compression (SNC)~\cite{SNC2014} is a leading sample compression method in high-dimensional input feature space. In contrast, SNC can only achieve up to 136 times speedup over kNN with comparable performance on MNIST with at least 600 learned exemplars~\cite{SNC2014}. That is, it only achieves a compression ratio as high as 30 times of that of en-HOPE. Part of the reason here is that it is not designed for data embedding and visualization and thus unable to compress dataset from the aspect of dimensionality reduction. This assumption is further verified by the following experimental observations. When using 20 learned exemplars in the high-dimensional input feature space, SNC produced test error rates of 6.31\% on MNIST and 17.50\% on USPS, which are much higher than those of en-HOPE. Also, if we pre-project data to two-dimensional space by other methods such as PCA or LMNN and then run SNC, the results of SNC should be much worse than the ones in the high-dimensional input feature space. Although the focus of en-HOPE is not for sample compression but for data embedding and visualization by collapsing classes, when we embed MNIST data to a 10-dimensional latent space using en-HOPE with 20 exemplars, we can further reduce the test error rate from $2.66$\% to $2.31$\%.

We also further evaluate the performance of our shallow model en-HOPE with 20 learned exemplars against deep method dt-MCML on the MNIST data. When compared to dt-MCML, en-HOPE achieves 316 times speedup for classifying MNIST test data in 2D owing to its proposed exemplar learning functionality. It is also worth mentioning that, although both methods have the overhead of computing the 2D embedding of test data, en-HOPE has 2 times speedup over dt-MCML on this burden owing to its shallow architecture. 

\section{Conclusion and Future Work}
\label{sec:discussion}
In this paper, we present an exemplar-centered supervised shallow parametric data embedding model en-HOPE by collapsing classes for data visualization and fast kNN classification. 
Owing to the benefit of a small number of precomputed or learned exemplars, en-HOPE avoids pairwise training data comparisons and only has linear computational cost for both training and testing.  Experimental results demonstrate that en-HOPE accelerates kNN classification by hundreds of times, outperforms state-of-the-art supervised embedding methods, and effectively collapses classes for impressive two-dimensional data visualizations in terms of both classification performance and visual effects. 

In the future, we aim to extend our method to an unsupervised learning setting to increase the scalability of traditional t-SNE, for which we just need to compute the pairwise probability $p(j|i)$ using high-dimensional feature vectors instead  of class  labels and optimize exemplars accordingly.

\bibliographystyle{named}
\bibliography{ref}

\end{document}